\documentclass[letterpaper]{article} 
\usepackage{aaai24}  
\usepackage{times}  
\usepackage{helvet}  
\usepackage{courier}  
\usepackage[hyphens]{url}  
\usepackage{graphicx} 
\usepackage{natbib}  
\usepackage{caption} 
\usepackage{algorithm}
\usepackage{algorithmic}

\usepackage{newfloat}
\usepackage{listings}
\DeclareCaptionStyle{ruled}{labelfont=normalfont,labelsep=colon,strut=off} 
\floatstyle{ruled}
\newfloat{listing}{tb}{lst}{}
\floatname{listing}{Listing}
\usepackage{graphicx}
\usepackage{amsmath}
\usepackage{amssymb}
\usepackage{booktabs}
\usepackage{multirow}
\usepackage{colortbl}
\usepackage{algorithm}
\usepackage{algorithmic}
\usepackage{xcolor}
\definecolor{mypink}{rgb}{.99,.91,.95}
\definecolor{mygreen}{RGB}{107,147,147}
\definecolor{lgreen}{HTML}{00b8a9}
\definecolor{mygray}{HTML}{FFFFFF}
\usepackage{arydshln}
\usepackage{xcolor}
\usepackage{color,colortbl}
    \definecolor{mypink}{rgb}{.99,.91,.95}

\usepackage{xcolor}
\usepackage{soul}
\usepackage{pifont}
\usepackage{leftidx}
\usepackage{varwidth}
\usepackage{capt-of}

\definecolor{darkergreen}{RGB}{21, 152, 56}
\newcommand\greenp[1]{{(#1)}}

\newcommand{\rr}[1]{{\textbf{#1}}}
\newcommand{\bb}[1]{{\underline{#1}}}

\usepackage{caption}
\usepackage{subcaption}

\makeatletter
\newcolumntype{I}{!{\vrule width 1.2pt}}
\newlength\savedwidth
\newcommand\whline{\noalign{\global\savedwidth\arrayrulewidth
                            \global\arrayrulewidth 1.5pt}
                   \hline
                   \noalign{\global\arrayrulewidth\savedwidth}}
\newlength\savewidth

\definecolor{darkergreen}{RGB}{21, 152, 56}

\usepackage{array} 
 
\title{Advancing Volumetric Medical Image Segmentation via Global-Local Masked Autoencoder}
\author{
   Jia-Xin Zhuang,  Luyang Luo, Hao Chen
}
\affiliations{
    Department of Computer Science and Engineering\\
  Hong Kong University of Science and Technology\
}

\usepackage{bibentry}

\begin{document}

\maketitle

\begin{abstract}
Masked autoencoder (MAE) is a promising self-supervised pre-training technique that can improve the representation learning of a neural network without human intervention. However, applying MAE directly to volumetric medical images poses two challenges: (i) a lack of global information that is crucial for understanding the clinical context of the holistic data, (ii) no guarantee of stabilizing the representations learned from randomly masked inputs. To address these limitations, we propose the \textbf{G}lobal-\textbf{L}ocal \textbf{M}asked \textbf{A}uto\textbf{E}ncoder (GL-MAE), a simple yet effective self-supervised pre-training strategy. In addition to reconstructing masked local views, as in previous methods, GL-MAE incorporates global context learning by reconstructing masked global views. Furthermore, a complete global view is integrated as an anchor to guide the reconstruction and stabilize the learning process through global-to-global consistency learning and global-to-local consistency learning.
Finetuning results on multiple datasets demonstrate the superiority of our method over other state-of-the-art self-supervised algorithms, highlighting its effectiveness on versatile volumetric medical image segmentation tasks, even when annotations are scarce. 
Our codes and models will be released upon acceptance.
\end{abstract}

\section{Introduction}
\label{sec:introduction}

Deep learning has shown great promise in medical image analysis, yet is limited to supervised learning based on a relatively large amount of labeled training data~\cite{shen2017deep}.
However, labeling medical images, especially volumetric images, can be expertise-dependent, labor-intensive, and time-consuming, which motivates remarkable progress in label-efficient learning~\cite{jin2023label,tajbakhsh2020embracing}.
Particularly, to enlarge the training set while keeping less human intervention, Self-Supervised Learning (SSL) approaches~\cite{jing2020self} have demonstrated their effectiveness by firstly pre-training on a large number of unlabeled data and then finetuning on small-scale labeled datasets to improve the task performance of the small-scale labeled dataset.
\begin{figure}[ht]
    \vspace{-15pt}
    \begin{center}
	\includegraphics[width=0.92\linewidth]{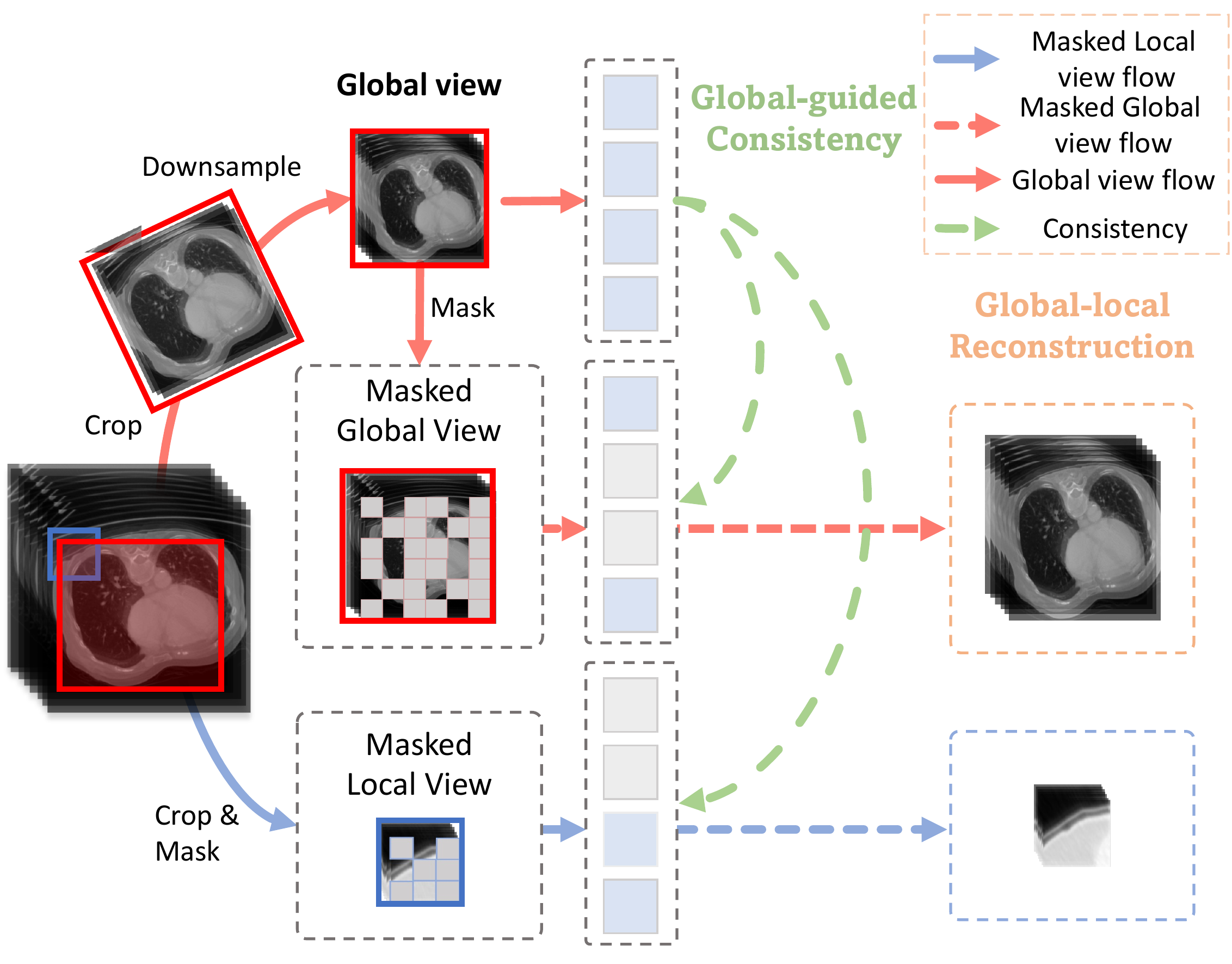}
    \end{center}
    \vspace{-10pt}
	\caption{Illustration of our proposed method. When utilizing a single volume image as input, MAE3D method directly employs a local masked volume reconstruction technique (represented by a blue dashed line). In contrast, our approach incorporates both masked global and local sub-volumes by utilizing global context guidance (Best view in colors).}\label{fig:highlight}
    \vspace{-15pt}
\end{figure}

In specific, SSL-based pre-training neglects the process of obtaining human annotations by learning representations with supervision signals generated from the data itself, which has become an important label-efficient solution for volumetric medical image representation learning \cite{chen2019self,hatamizadeh2022unetr,tang2022self,tao2020revisiting,zhou2023unified}.

Recently, Vision Transformer (ViT)~\cite{dosovitskiy2020image} has inspired an increasing number of SSL approaches due to its high scalability and generalization ability.
For example, Masked Autoencoder (MAE)~\cite{he2022masked} and SimMIM~\cite{xie2022simmim} have achieved high performance in various natural image analysis tasks by learning transferable representations via reconstructing the original input in pixel from its highly-masked version based on the ViT structure.
However, for medical images such as Computed Tomography (CT), which are often volumetric and large in size, existing methods like Swin-UNETR \cite{tang2022self} and MAE3D \cite{chen2023masked} have proposed breaking down the original volume scans into smaller sub-volumes (e.g., a $96\times 96\times 96$ sub-volume from a $512\times 512\times 128$ CT scan) to reduce the computation cost of ViT.

However, the use of local cropping strategies for medical images poses two significant challenges. Firstly, these strategies focus on reconstructing information from the masked \textit{local} sub-volumes, neglecting the \textit{global} context information of the patient as a whole. Global representation has been shown to play a crucial role in Self-Supervised Learning \cite{caron2021emerging,xu2021self,zhang2022leverage,fan2022self}. For medical images, a global view of the volumetric data, as shown in Figure~\rr{\ref{fig:highlight}}, contains rich clinical context of the patient, such as the status of other organs, which is essential for further analysis and provides important clinical insights. Secondly, there is no guarantee that the learned representations will be stable to the input distortion caused by masking, particularly when the diverse local sub-volumes only represent a small portion of the original input. This can lead to slow convergence and low efficiency during training. Pre-training solely with strong augmentation, such as a local view masked with a high ratio, is considered a challenging pretext task, but it may distort the image structure and result in slow convergence. Instead, weak augmentation can be seen as a more reliable "anchor" to the strong augmentations~\cite{zheng2021ressl,wang2022contrastive}.

To address the challenges mentioned above, we propose a straightforward yet effective SSL approach called Global-Local Masked AutoEncoder (GL-MAE). As shown in Figure~\rr{\ref{fig:highlight}}, we obtains both the global view and the local view of an input volumetric image by applying image transforms such as cropping and downsampling for the global views. On the one hand, the global view covers a large region with rich information but low spatial resolution, which may miss details for small organs or tumors. On the other hand, the local views are rich in details but only take up a small fraction of the input volume and in high spatial resolution. To leverage both sources of information, we propose to use an MAE to simultaneously reconstruct both the masked global and local images, enabling learning from both the global context and the local details of the data. Moreover, we noticed that a global view of the image encompasses a more comprehensive area of the data and could help learn representations that are invariant to different views of the same object~\cite{zhang2022leverage}. Such view-invariant representations are beneficial in medical image analysis. To encourage the learning of view-invariant representations, we propose global-guided consistency learning, where the representation of an unmasked global view is used to guide learning robust representations of the masked global and local views. Finally, as the global view covers most of the local views, it can serve as an "anchor" for the masked local views to learn the global-to-local consistency. 

In summary, we present GL-MAE, an MAE-based SSL algorithm for learning representations of volumetric medical data. 
GL-MAE involves reconstructing input volumes from both global and local views of the data.
It also employs consistency learning to enhance the semantic corresponding by using unmasked global sub-volumes to guide the learning of multiple masked local and global views. 
By introducing the global information into the MAE-based SSL pre-training, GL-MAE achieved superior performance in the downstream volumetric medical image segmentation tasks compared with  state-of-the-art MAE3D and Swin-UNETR.
\section{Related Works}\label{sec:relatedWork}

\textbf{Self-Supervised Learning with Medical Image Analysis.} Self-Supervised Learning (SSL) is an unsupervised method that learns representations for neural networks with supervision signals generated from the data itself. Previous studies in SSL can be categorized into three primary paradigms: contrastive-based, pretext task-based, and clustering-based methods~\cite{gao2022disco}. 

\textit{Contrastive learning } methods bring positive pairs of images (e.g., different views of the same image) closer and separate negative pairs (e.g., different images) away in the feature space~\cite{chen2020exploring,chen2020big,chen2020simple,he2019moco,chen2020mocov2}.
For example, MoCo-V2 utilized a memory bank to maintain consistent representations of negative samples during contrastive learning~\cite{chen2020improved}. DeSD~\cite{ye2022desd} proposed to use the feature from the deep layers of a teacher model to supervise the learning process of the shallow layers of a student model.
These works have also shown effectiveness in medical image analysis, such as dermatology classification and chest X-ray classification~\cite{azizi2021big,zhao2021unsupervised,zhou2021preservational}.  
However, previous contrastive learning approaches have primarily focused on the semantic misalignment of different instances, leading to minor improvements in dense prediction, such as segmentation~\cite{chaitanya2020contrastive}.

\textit{Pretext task-based} methods, on the other hand, are designed to explore the inner space structure information of images and have shown promise in dense prediction tasks~\cite{chen2019self,haghighi2022dira,tao2020revisiting,xie2020pgl,zheng2021hierarchical}. For example, Model Genesis proposed to directly pre-train a CNN model by restoring volumetric medical images from their distorted versions~\cite{zhou2021models}. Rubik’s Cube+ utilized the concept of solving a Rubik’s Cube to learn structural features from the volumetric medical data~\cite{zhu2020rubik}. Additionally, five common pretext tasks have been verified to be effective for volumetric image pre-training~\cite{taleb20203d}. Swin-UNETR~\cite{tang2022self} proposed to pretrain a 3D swin transformer \cite{liu2021swin} with the combination of three pretext tasks, including inpainting, contrastive learning, and rotation prediction. 

\begin{figure*}[!ht]
	\includegraphics[width=1.0\linewidth]{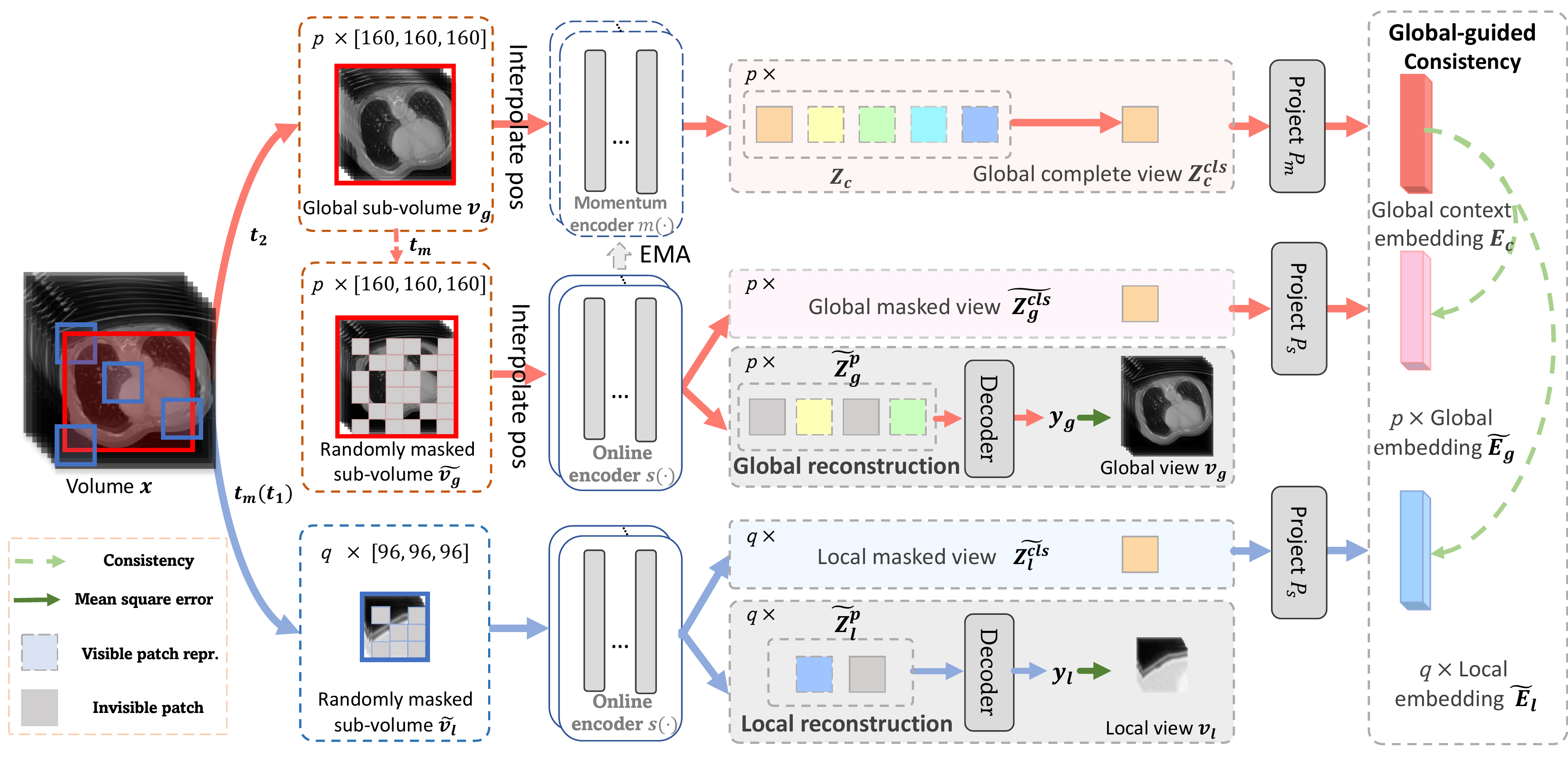}
	\vspace{-15pt}
	\caption{The framework of our method. Two types of sub-volumes, $p$ global view $v_g$ and the $q$ local views $v_l$, are extracted from a 3D volumetric data (e.g. a CT image scan), respectively. The obtained sub-volumes are then masked to be $p \widetilde{v_g}$ and $q \widetilde{v_l}$ prior to being partitioned, encoded, and reconstructed at different scale by the vision transformer. Meanwhile, the representation of the unmasked global sub-volume $v_g$ is used to guide the learning of the representations of the masked local volumes $\widetilde{v_l}$ as well as the masked global volumes $\widetilde{v_g}$. More samples for different views can be found in the supplementary material.
 } 
	\label{fig:framework}
    \vspace{-10pt}
\end{figure*}
\noindent\textbf{Mask Image Modeling.} Masked image modeling is a generative SSL technique that learns feature representations by training on images that are corrupted by masking. 
Early research on masked image modelings, such as DAE~\cite{vincent2010stacked} and context encoder~\cite{pathak2016context}, treated masking as a type of noise and used inpainting techniques to predict the missing pixels. 
ViT~\cite{dosovitskiy2020image} investigated masked patch prediction by predicting the mean color of images. MAE~\cite{he2022masked} adhered to the spirit of raw pixel restoration and demonstrated for the first time that masking a high proportion of input images can yield a non-trivial and meaningful self-supervisory task. SimMIM~\cite{xie2022simmim} took this approach one step further by substituting the entire decoder with a single linear projection layer, resulting in comparable results. MAE3D~\cite{chen2023masked} represented a recent advancement in masked image modeling for 3D CT in medical image analysis. 
This method proposed to recover invisible patches from visible patches using a transformer backbone with a high mask ratio and achieved competitive results. However, it should be noted that such methods for 3D CT in medical image analysis were based on local cropping strategies which did not consider global information in the training process.

\section{Method}\label{sec:method}

GL-MAE aims to address the aforementioned challenges of existing ViT-based volumetric medical image pre-training.
In a nutshell, GL-MAE mainly consists of two parts: a masked autoencoder with Global and Local Reconstruction, and Global-guided Consistency Learning. First, reconstructing both the global views and local views enables MAE model to learn the global context as well as the local details.
Then, consistency learning is introduced between the representations of unmasked global views and the masked views, which encourages invariant and robust feature learning.

\subsection{MAE with Global and Local Reconstruction}\label{sec:reconstruction} 
The framework involves an unlabelled CT dataset $\mathcal{D}=\{x_1, x_2,... x_N\}$, where $N$ is the total number of volumes. As shown in Figure~\ref{fig:framework}, a volume $x \in \mathbb{R}^{CHWD}$ is randomly sampled from $\mathcal{D}$, and then augmented into a small-scale sub-volume $v_l\in \mathbb{R}^{CHWD}$ using image transformation $\tau_1$ and a large-scale sub-volume $v_g\in \mathbb{R}^{CHWD}$ using image transformation $\tau_2$. The local view $v_l$ provides detailed information in high resolution about texture and boundaries, but lacks a global view of the input data, such as the status of other organs. On the other hand, the global view $v_g$ provides a macro-level view of a large area within the volume, but in low resolution, which is important for downstream tasks involving dense prediction. This process is repeated $q$ times for local sub-volumes and $p$ times for global sub-volumes to form a local view's set $V_l=\{v_l^i, i\in [1, q]\}$ and a global view's set $V_g=\{v_g^i, i\in [1, p]\}$. 

To obtain the masked volumes for local and global reconstruction, all local views $V_l$ and global views $V_g$ will be individually tokenized into patches and then applied volume masking transform $\tau_m$ with a pre-defined ratio. These masked patches will serve as the invisible patches, while the rest of the patches will serve as input for the learnable encoder $s(\cdot)$. By applying $\tau_m$ for each $v_g$ and $v_l$ respectively, the visible patches forms a set of masked local and global sub-volumes $\widetilde{V_l}=\{\widetilde{v_l^i}, i\in [1, q]\}$ and $\widetilde{V_g}=\{\widetilde{v_g^i}, i\in [1, p]\}$.

The encoder $s(\cdot)$  is used to map the input volumes to a representation space. The position embedding is added to each visible patch and then combined with a class token to generate the volume representation. These visible patches for local and global views $v_l$ and $v_g$ will be fed into $s(\cdot)$ to generate $\widetilde{Z_l}$ and $\widetilde{Z_g}$ by:
\begin{equation}\label{eq:z_embed}
    \widetilde{Z_l}=s(\widetilde{v_l}),\quad \widetilde{Z_g}=s(\widetilde{v_g}),
\end{equation}
where $Z$ consists of two part embeddings, including the output of the class token and patch tokens, denoted as $\widetilde{Z_l}\triangleq[\widetilde{Z_l^{cls}};\widetilde{Z_l^{p}}]$ , $\widetilde{Z_g}\triangleq[\widetilde{Z_g^{cls}};\widetilde{Z_g^{p}}]$
. $\widetilde{Z_g^{cls}}$ and $\widetilde{Z_l^{cls}}$ are outputs of the class tokens, while $\widetilde{Z_g^{p}}$ and $\widetilde{Z_l^{p}}$ are outputs of the visible patches, which then are used for reconstruction.

A decoder $\mathcal{D}(\cdot)$ used to reconstruct the invisible patches from the representation of the visible patches. The input of the decoder $\mathcal{D}(\cdot)$ consists of (i). encoded visible patches and (ii). masked token. As shown in Figure~\textbf{\ref{fig:framework}}, the mask token is learnable and indicates the missing patches to predict. Position embedding to all tokens are added to cover the location information. The output of the decoder $\mathcal{D}(\cdot)$ can derive by:
\begin{equation}\label{eq:y}
     y_l=\mathcal{D}({\widetilde{Z_l^{p}})}\quad,
     y_g=\mathcal{D}({\widetilde{Z_g^{p}})},
\end{equation}
Decoder output is reshaped to form reconstructed volumes. Mean Square Error is used as reconstruction loss function and applied to the local and global masked sub-volumes.

\noindent\textbf{Local reconstruction.} For local masked sub-volumes $\widetilde{V_l}$, the local reconstruction loss $\mathcal{L}_\mathcal{R}^l$ can be defined as:
\begin{equation}
    \mathcal{L}_\mathcal{R}^l = \frac{1}{{|\widetilde{V_l}|\times HWD}}\sum_{h=1}^{H}\sum_{w=1}^{W}\sum_{d=1}^{D}\left(\sum_{v_l\in \widetilde{V_l}}\frac{(y^{h,w,d}_l-v^{h,w,d}_l)^2}{P^{}}\right),
	\label{eq:reconstructionLossLocal}
\end{equation}
where ${h,w,d}$ denotes voxel indices on the representations, $P$ represent the numbers of patches for local views, while $H, W, D$ refer to the height, width, and depth of each sub-volume, respectively. The reconstruction loss is computed as the sum of squared differences between the reconstruction target and the reconstructed representations by pixel values.

\noindent\textbf{Global reconstruction.} Similarly, for global masked sub-volumes $\widetilde{V_g}$, the global reconstruction loss $\mathcal{L}_\mathcal{R}^g$ defined as:
\begin{equation}
    \mathcal{L}_\mathcal{R}^g = \frac{1}{|{\widetilde{V_g}|\times HWD}}\sum_{h=1}^{H}\sum_{w=1}^{W}\sum_{d=1}^{D}\left(\sum_{v_g\in \widetilde{V_g}}\frac{(y^{h,w,d}_g-v^{h,w,d}_g)^2}{P^{'}}\right),
	\label{eq:reconstructionLossGlobal}
\end{equation}
where $P^{'}$ represent the numbers of patches for global views. It's notes that since global view has a larger input size than local views, position embedding needs to be interpolated before being added to the visible tokens. This process enables the reconstruction of the masked volumes at both the local and global views, facilitating the learning of rich information from both the local details and global information.

\subsection{Global-guided Consistency Learning}
The global-guided consistency learning approach includes two components: global-to-global consistency and global-to-local consistency. The first component enforces consistency between the representations of the unmasked global view $v_g$ and the masked global views $\widetilde{v_g}$, promoting the learning of features that are robust to the distortion caused by masking and accelerating training convergence. Since the global view contains richer context and covers most of the local views, it can be used as an "anchor" to guide the representation learning of the local views~\cite{zhang2022leverage}. The second component, global-to-local consistency is crucial for capturing information about the relationships between different parts of an image and its main semantics~\cite{caron2021emerging}. It enforces consistency between representations of the unmasked global view $v_g$ and the masked local view $\widetilde{v_l}$. 

As depicted in Figure~\rr{\ref{fig:framework}}, the global complete view $Z_c^{cls}$ guides the representation learning of the global masked volumes $\widetilde{Z_g^{cls}}$ and local masked volumes $\widetilde{Z_l^{cls}}$. To obtain $\widetilde{Z_c^{cls}}$, a two-encoder architecture is used, consisting of a learnable encoder $s(\cdot)$ based on the transformer and a momentum encoder $m(\cdot)$. The learnable encoder $s(\cdot)$ focuses on learning features from the masked views, while the momentum encoder $m(\cdot)$ generates the mean representation of the unmasked global view $v_g$ as $Z_c = m(v_g), v_g\in V_g$. The momentum encoder's parameters are updated using a momentum factor that is dynamically computed based on the learnable encoder's parameters as follows:
\begin{equation}\label{eq:momentum_update}
m^{(t)}(\cdot) \leftarrow \mu s^{(t)}(\cdot)+(1-\mu) m^{(t-1)}(\cdot),
\end{equation}
\noindent where $m^{(t)}(\cdot)$ and $s^{(t)}(\cdot)$ represents the momentum encoder and encoder at the $t$-th iteration, respectively, and $\mu$ is the momentum coefficient updated with a cosine scheduler. 

To perform consistency learning, the representations of the global complete view $Z_c^{cls}$ and the masked views $\widetilde{Z_g^{cls}}$, $\widetilde{Z_l^{cls}}$ must first be projected into the shared space. Projection layers $\mathcal{P}_s(\cdot)$ and $\mathcal{P}_m(\cdot)$   are introduced and follow the encoder $s(\cdot)$ and $m(\cdot)$, respectively. These projections are implemented using multiple fully-connected layers followed by a Gaussian Error Linear Unit activation function \cite{hendrycks2016gaussian}. After projection, the embedding of complete global views $E_c$, the masked global views $\widetilde{E_g}$, and the masked local view $\widetilde{E_l}$ are defined as below:
\begin{equation}\label{eq:embed_f}
    E_c = \mathcal{P}_m(Z_c^{cls}), \widetilde{E_g} = \mathcal{P}_s(\widetilde{Z_g^{cls}}), 
     \widetilde{E_l} = \mathcal{P}_s(\widetilde{Z_l^{cls}}).
\end{equation}
The dimension of $E_c$, $\widetilde{E_g}$ and $\widetilde{E_l}$ are $K$, e.g., 512. For each type of embedding $E$ in the shared space, they are normalized before computing the loss function. The embedding  $E^{i}$ is normalized as follows:
\begin{equation}\label{eq:normalize}
    \Gamma{(E^{i})} =\frac{\exp \left(E^{i} / t\right)}{\sum_{k=1}^K \exp \left(E^{k} / t\right)}, 
\end{equation}
where $t$ represent temperature for control the entropy of the distribution. Our goal is to minimize distributions between the representations of the global complete view $E_c$ and the masked global view $\widetilde{E_g}$ as well as the masked local view $\widetilde{E_l}$ via cross-entropy loss $H(x, y)=-x{\rm log} y$ as follows:
\begin{equation}\label{consistent_minimize}
    H\left(\Gamma({E_c}), \Gamma({\widetilde{E_g}})\right) + H\left(\Gamma({E_c}), \Gamma({\widetilde{E_l}})\right).
 \end{equation}

\noindent{\textbf{Global-to-global consistency.}} For global unmasked sub-volumes $V_g$ and global masked sub-volumes $\widetilde{V_g}$. The global-to-global consistency loss function can be formulated as:
\begin{equation}
    \mathcal{L}_\mathcal{C}^{gg} = \frac{1}{|V_g|\cdot|\widetilde{V_g}|}\left\{
    \sum_{v_g \in V_g} \sum_{\widetilde{v_g} \in \widetilde{V_g}} H\left(\Gamma({E_c}),\Gamma({\widetilde{E_g}})\right)\right\}, 
	\label{eq:constrastiveLossGlobal}
\end{equation}
where $|\cdot|$ computes the number of volumes in the set.  
$\widetilde{E_g}$ learns consistency guided by global context embedding $E_c$.

\noindent{\textbf{Global-to-local consistency.}}
Similarly, for global unmasked sub-volumes $V_g$ and local masked volumes $\widetilde{V_l}$, global-to-local consistency loss function can be obtained by:
\begin{equation}
    \mathcal{L}_\mathcal{C}^{gl} = \frac{1}{|V_g|\cdot|\widetilde{V_l}|}\left\{\sum_{v_g \in V_g} \sum_{\widetilde{v_l} \in \widetilde{V_l}} H\left(\Gamma({E_c}), \Gamma({\widetilde{E_l}})\right)\right\}. 
	\label{eq:constrastiveLossLocal}
\end{equation}
Local embedding $\widetilde{E_l}$ learn consistency guided by global context embedding $E_c$ during the pre-training process. 

\subsection{Overall Objective}

\begin{table*}[!ht]
    \centering
    \caption{Comparison of our proposed method using ViT-B as the backbone with SOTA approaches on transfer learning to three datasets under different ratios of training datasets. SL represents Supervised Learning. Results are evaluated using the Dice score (\%) metric and all methods were trained and evaluated on the same split. Supervised baselines without pretraining are included for comparison. Best and second-best results are highlighted in bold and underline, respectively. Pretrained checkpoints were obtained from the official implementation.}\label{tab:sota}
    \vspace{-8pt}
    \resizebox{1.0\linewidth}{!}{
    \begin{tabular}{c|c|c|ccc|ccc|ccc}
    \whline
        \multirow{2}{*}{Paradigm} & \multirow{2}{*}{Method} & \multirow{2}{*}{Conference} & \multicolumn{3}{c|}{$\rightarrow$BTCV} & \multicolumn{3}{c|}{$\rightarrow$MSD Spleen} & \multicolumn{3}{c}{$\rightarrow$MM-WHS}\\ 
        \cline{4-12}
        & & ~ & 25\% & 50\% & 100\% & 25\% & 50\% & 100\% & 25\% & 50\% & 100\% \\ 
        \hline
        \multirow{3}{*}{SL} & 3D U-Net~\cite{ronneberger2015u} & MICCAI 2015 & 59.45  & 73.01  & 79.53  & 80.28  & 91.68 & 93.71 & 66.19  & 79.52  & 83.09  \\ 
        & Segresnet~\cite{myronenko20193d} & MICCAIW 2016 & 31.05  & 71.42  & 79.97  & 78.32  & 91.85 & 94.10 & 63.62  & 80.94  & 83.25 \\ 
        \cline{2-12}
        & UNETR & WACV 2022 & 58.99  & 73.17  & 79.61  & 84.57  & 92.23 & 94.20 & 66.14  & 80.10  & 83.85  \\ 
        \hline
        \multirow{7}{*}{SSL} & ModelGen~\cite{zhou2021models} & MIA 2021 & 54.18  & 61.26 & 81.45  & 85.65  & 93.02 & 94.40 & 72.28  & 81.33  & 86.36  \\ 
        & VicRegl~\cite{bardes2022vicregl} & NeurIPS 2022 & - & - & - & -& - & - & - & - & 84.72\\
        & Swin-UNETR~\cite{tang2022self} & CVPR 2022 & {63.96}  & 75.15  & 81.54  & 88.56  & 93.54 & 95.02 & 70.74  & 81.66  & 87.06  \\ 
        & JSSL~\cite{nguyen2023joint} & AAAI 2023 & - & - & - & -& - & - & - & - & 84.89\\
        & GLSV~\cite{he2023geometric} & CVPR 2023 & 40.20 & 46.84 & 55.72 & 89.05  & \bb{94.20} & \bb{95.47}  & \bb{74.51}  & \rr{85.08}  & \bb{87.07}  \\ 
        & MAE3D~\cite{chen2023masked} & WACV 2023 & \bb{66.08}  & \bb{75.39}  & \bb{81.74} & \bb{89.65}  & \bb{94.20} & 95.20 & 69.78  & \bb{84.64}  & 86.03  \\ 
        & GL-MAE (Ours) & - & \rr{66.44}  & \rr{76.37}  & \rr{82.33}  & \rr{90.65}  & \rr{94.36} & \rr{95.72} & \rr{76.16}  & 83.72 & \rr{88.88}  \\ 
        \whline
    \end{tabular}}
    \vspace{-10pt}
\end{table*}
The overall objective function is represented by 
\begin{equation}
    \mathcal{L} = \mathcal{L}_{R}^{l} + \beta_1\mathcal{L}_{R}^{g}+ \beta_2\mathcal{L}_{C}^{gg}+\beta_3\mathcal{L}_{C}^{gl},
\end{equation}
where the hyper-parameters $\beta_1$, $\beta_2$, and $\beta_3$ are used to balance the relative contributions of these four loss terms and set to 1.0 in experiments empirically. A pseudocode of the overall framework is shown in the supplementary material.


\section{Experiments}
\label{sec:exp}

\noindent\textbf{Datasets and evaluation metrics.} The SSL pretraining experiments were carried out on the \textit{Beyond the Cranial Vault (BTCV)} abdomen challenge dataset~\cite{landman2015miccai} and \textit{TCIA Covid19 dataset}~\cite{an2020ct}. For downstream tasks, experiments were mainly conducted on the BTCV dataset to follow previous work~\cite{chen2023masked,tang2022self}. To assess the model's generalization on Computed Tomography (CT) datasets, we also evaluated its effectiveness on \textit{MM-WHS}~\cite{zhuang2018multivariate}, \textit{Medical Segmentation Decathlon (MSD) Task 09 Spleen}, and \textit{The COVID-19-20 Lung CT Lesion Segmentation Challenge dataset (Covid-19-20 dataset)}~\cite{roth2022rapid}. The model was further transferred to \textit{Brain Tumor Segmentation (BrasTS)}~\cite{simpson2019large} for assessing its cross-modality generalization ability. All datasets used were collected from \textbf{\textit{open source}} and can be obtained via the cited papers. More details of the datasets can be found in Section Datasets in the supplementary material.
Dice Score(\%) was used as the evaluation metric following~\cite{chen2023masked,tang2022self}.


\noindent\textbf{Pre-training setting.}  ViT~\cite{dosovitskiy2020image} is a well-known and strong transformer-based backbone. Specifically, we have used both ViT-Tiny (ViT-T) and ViT-Base (ViT-B) for our experiments. The pretraining phase was conducted for 1600 epochs for ViT-T and ViT-B without specification, with an initial learning rate of 1e-2, employing AdamW~\cite{kingma2014adam} as an optimizer and a batch size of 256 on four 3090Ti for 3 days. For global views $v_g$, images were scaled by a random ratio from the range of [0.5, 1], cropped, and then resized into [160, 160, 160], while for local views $v_l$, images were scaled by a random ratio from the range of [0.25, 0.5], cropped and resized into [96, 96, 96]. Finally, all images were normalized from [-1000, 1000] to [0,1]. $p$ and $q$ set to 2 and 8.
More details on transformations can be found in the supplementary material in Section Implementation details.

\noindent{\textbf{Finetuning setting.} UNETR~\cite{hatamizadeh2022unetr} is adopted as the segmentation framework. We introduce details for finetuning on BTCV dataset here. For linear evaluation that freezes the encoder parameters and finetuning the segmentation decoder head, model was finetuned for 3000 epochs using an initial learning rate of 1e-2, and trained on a single 3090Ti GPU with a batch size of 4. For end-to-end segmentation, the model was trained on four 3090Ti GPUs for 3000 epochs, with a batch size of 4, using an initial learning rate of 3e-4. More information about finetuning on other datasets can be found in Section Implementation details in the supplementary material.

\begin{table*}[!ht]
    \centering
    \caption{Comparison with SOTA on BTCV validation dataset under the Linear and End-to-end evaluation, with ViT-T as the backbone for all the methods. Linear evaluation indicates fixing the parameters of encoder and fine-tuning the rest blocks. Hausdorff Distance-95 calculates 95th percentile of surface distances between ground truth and prediction point sets. }
    \vspace{-8pt}
    \resizebox{0.92\linewidth}{!}{
    \begin{tabular}{c|ccc|ccc|ccc}
    \whline
        Metric & \multicolumn{3}{c|}{Dice score(\%)$\uparrow$} & \multicolumn{3}{c|}{Normalized surface dice(\%)$\uparrow$} & \multicolumn{3}{c}{Hausdorff distance-95$\downarrow$} \\ \hline
        Setting & Baseline & MAE3D & GL-MAE & Baseline & MAE3D & GL-MAE & Baseline & MAE3D & GL-MAE \\ \hline
        Linear & - & 78.36 & 80.22 & - & 38.79 & 41.47 & - & 16.41 & 13.03 \\ 
        End-to-end & 78.01 & 80.21 & \textbf{81.02}\greenp{3.01$\uparrow$} & 36.80 & 39.72 & \textbf{41.45}\greenp{4.65$\uparrow$} & 14.35 & 5.46 & 7.53\greenp{6.82$\downarrow$} \\ 
    \whline
    \end{tabular}}
    \label{tab:btcv_vitt}
    \vspace{-10pt}
\end{table*}
\subsection{Experiment results on downstream tasks}
\noindent\textbf{End-to-end and linear evaluation.}\label{sec:linearEvaluationEnd2End}
To assess the effectiveness of our proposed method, following~\cite{tang2022self,chen2023masked,he2023geometric}, we conducted end-to-end segmentation experiments on three datasets . 
In Table~\rr{\ref{tab:sota}}, our proposed method GL-MAE (10th row) outperformed the Supervised baseline (3rd row) by a large margin (82.33\% vs 79.61\%, 95.72\% vs 94.20\%, and 88.88\% vs 83.85\%) with full training dataset, indicating that our method GL-MAE benefits the model from the unlabeled dataset. MAE3D is a recently proposed competitive SSL strategy in medical image analysis, where it has shown superiority (9th row), particularly on dense prediction tasks such as segmentation. Our proposed method GL-MAE (10th row) outperformed the MAE3D (9th row), which further confirms the effectiveness of our approach. SegresNet and 3D U-Net are supervised models with competitive performance. Swin-UNETR uses 3D Swin-Transformer as the backbone, while GLSV was designed for Cardiac CT images. All SSL methods (4-10th row) achieved better performance than the supervised Baseline, while our method GL-MAE (10th row) achieved the best performance over three datasets even when using only 25\% and 50\% annotations of the training datasets. This indicates the superior generalization ability of GL-MAE.

GL-MAE showed consistent performance when using a more lightweight transformer, ViT-T, which requires less computational resources and can be trained and inferred faster. In Table~\rr{\ref{tab:btcv_vitt}} our method outperformed other methods by a large margin in terms of average Dice score, Normalized surface dice, and Hausdorff distance metric in both linear and end-to-end segmentation evaluation settings. This demonstrates the versatility of GL-MAE when adapting to a lightweight backbone, which is necessary in certain situations such as surgical robots.

\noindent\textbf{Generalization on the unseen datasets.} MM-WHS is a predominant small-scale organ dataset that has not been involved in the pretraining. As shown in Figure~\ref{fig:transfer_mmwhs}, the experimental findings demonstrated that the proposed GL-MAE significantly improved the average dice score from 86.03\% to 88.88\% compared with the MAE3D, indicating its strong generalization capabilities. Furthermore, there were substantial enhancements in the performance of the aorta, LV, and RV, which share analogous structural features with the training data. This suggests that our proposed method can exploit the structural consistency between organs across varying datasets and generalize effectively to the novel unseen datasets. 

\begin{figure}[ht]
	\centering
	\includegraphics[width=1.0\linewidth]{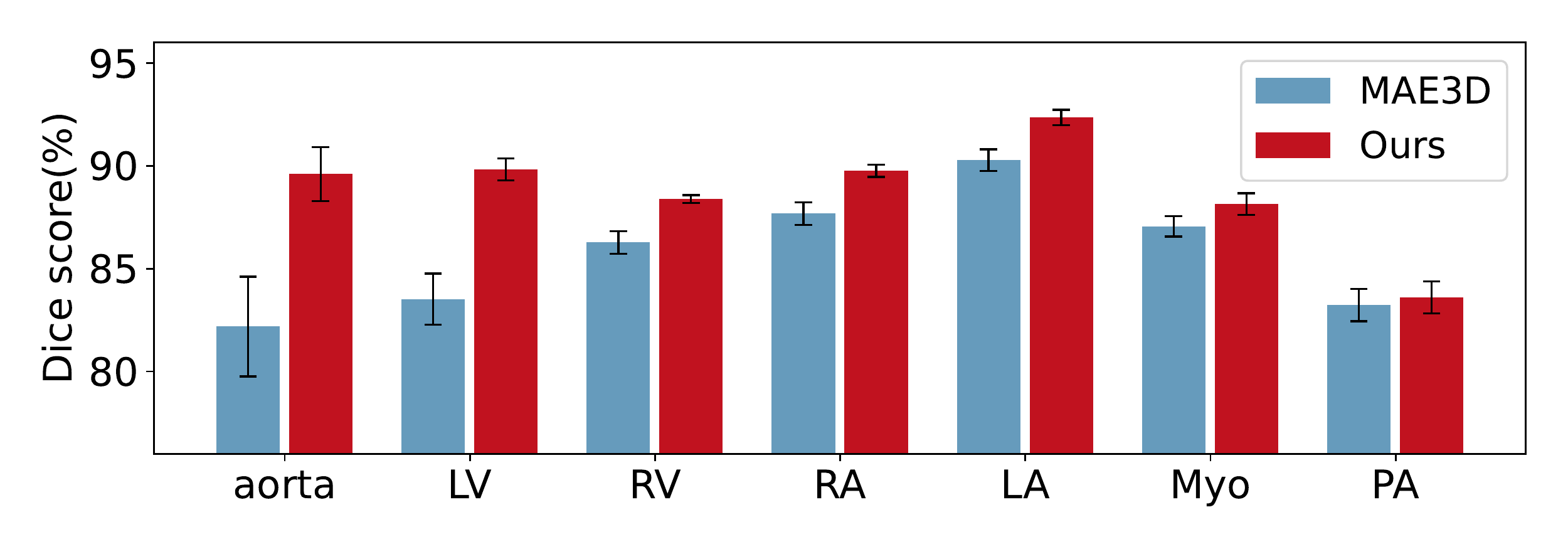}
    \vspace{-10pt}
	\caption{Dice score(\%) over each class respectively, of the segmentation performance on the MM-WHS datasets. Error calculation based on intra-organ comparisons across multiple cases. MM-WHS contains 7 classes including Left Ventricle (LV), whole aorta (aorta), Right Ventricle (RV), Left Atrium (LA), myocardium of LV (Myo), Right Atrium (RA), and Pulmonary Artery (PA).}
	\vspace{-5pt}
	\label{fig:transfer_mmwhs}
\end{figure}

\noindent\textbf{Generalization on Magnetic Resonance Imaging (MRI).}
We sought to evaluate whether our proposed method could enhance model performance on MRI datasets by pre-training on the CT datasets. Following~\cite{chen2023masked}, we conducted finetuning experiments on the Brain Tumor Segmentation MRI datasets (BraTS). As shown in Table~\rr{\ref{tab:transfer_msd}}, our proposed method improved model performance in both datasets despite the shift in modality, suggesting that similar organizational knowledge can be shared between different imaging modalities. 
As such, we remain curious as to whether the proposed method can be directly transferred to MRI datasets, and we plan to investigate this in future work.
\begin{table}[ht]
    \centering
    \caption{Dice score(\%) of brain tumor segmentation performance in BraTS. WT, ET, and TC denote Whole Tumor, Enhancing tumor, and Tumor Core sub-regions, respectively.}
	\vspace{-8pt}
    \resizebox{0.88\linewidth}{!}{
     \begin{tabular}{cccccc}
    	\whline
    	\multirow{2}{*}{Method} & \multirow{2}{*}{Backbone} & \multicolumn{4}{c}{Dice score(\%)}\\
    	\cline{3-6}  & & TC & WT & ET & \textbf{Avg}\\
    	\hline
    	\rowcolor{mygray}Baseline & ViT-T  & 81.62 & 87.81 & 57.34 & 75.59 \\
    	\cdashline{1-6}
        MAE3D & ViT-T & 82.34 & 90.35 & 59.18 & 77.29 \\
    	Ours & ViT-T & \textbf{83.00} & \textbf{91.15} & \textbf{60.46} & \textbf{78.31}\\
        \whline
    \end{tabular}}
    \label{tab:transfer_msd}
   \vspace{-5pt}
\end{table}

\noindent\textbf{COVID-19 lesion segmentation.}\label{sec:transferLesion}
CT scans are commonly used in diagnosing COVID-19, yet there is a shortage of annotated data. 
Our proposed method improves COVID-19 lesion segmentation performance from 47.92\% to 49.88\% compared with the baseline (please refer to Section Experiments in the supplementary material for detail). 
These findings suggest that our pre-trained model can capture valuable knowledge from unlabelled CT datasets to improve disease diagnosis, demonstrating the versatility of the proposed method in practical clinical settings.
\begin{table*}[ht]
    \centering
    \caption{Ablation study on the BTCV validation dataset and MM-WHS validation dataset. ViT-B and ViT-T were utilized as the backbone for the evaluation. The improvements achieved with this approach were compared to those of the supervised baseline. Sup. represents the supervised baseline. All models were pre-trained with 1000 epochs. Dice score(\%) was used as the metric.}
	\vspace{-8pt}
    \resizebox{0.95\textwidth}{!}{%
    \begin{tabular}{ccccc|cc|cc|cc}
            \whline
            \multicolumn{5}{c|}{Setting}  & \multicolumn{2}{c|}{Linear} & \multicolumn{2}{c|}{End-to-end} & \multicolumn{2}{c}{End-to-end}
            \\ \hline
            \multirow{2}{*}{Method} & \multicolumn{4}{c|}{\textbf{Loss}} & \multicolumn{2}{c|}{BTCV} & \multicolumn{2}{c|}{BTCV} & \multicolumn{2}{c}{MM-WHS}\\
            \cline{6-7} \cline{8-9} \cline{10-11}
            & $\mathcal{L}_\mathcal{R}^l$ & $\mathcal{L}_\mathcal{R}^g$ & $\mathcal{L}_\mathcal{C}^{gg}$ & $\mathcal{L}_\mathcal{C}^{gl}$ & ViT-B & ViT-T & ViT-B & ViT-T & ViT-B & ViT-T\\
            \hline 
             \rowcolor{mygray} Baseline & - & - & - & - & 79.04 & 77.18 & 79.61 & 78.02 & 83.85 & 83.04\\
            \cdashline{1-11}
            \multirow{4}{*}{GL-MAE} & \checkmark & & & & 79.32\greenp{0.28$\uparrow$} & 77.35\greenp{0.17$\uparrow$} & 81.74\greenp{2.13$\uparrow$} & 80.67\greenp{2.65$\uparrow$} & 86.58\greenp{2.73$\uparrow$} & 84.00\greenp{0.96$\uparrow$}\\
            & \checkmark & \checkmark & & & 79.79\greenp{0.75$\uparrow$} & 78.20\greenp{1.02$\uparrow$} & 81.77\greenp{2.16$\uparrow$} & 78.70\greenp{0.68$\uparrow$} & 86.77\greenp{2.92$\uparrow$} & 84.14\greenp{1.10$\uparrow$}\\
            & \checkmark & \checkmark & \checkmark & & 79.84\greenp{0.80$\uparrow$} & 78.82\greenp{1.64$\uparrow$} & 81.73\greenp{2.12$\uparrow$} & 80.42\greenp{2.40$\uparrow$} & 86.60\greenp{2.75$\uparrow$} & 84.39\greenp{1.35$\uparrow$}\\
            & \checkmark & \checkmark & \checkmark & \checkmark & 79.99\greenp{0.95$\uparrow$}
 & 78.93\greenp{1.75$\uparrow$} & 81.91\greenp{2.30$\uparrow$} & 80.76\greenp{2.74$\uparrow$} & 87.89\greenp{4.04$\uparrow$} & 84.56\greenp{1.52$\uparrow$}\\
            \whline
        \end{tabular}
        }
    \label{tab:ablation_study}
	\vspace{-5pt}
\end{table*}

\subsection{Analysis of our proposed framework}
\label{sec:abaltionStudy}

\noindent\textbf{Ablation study.} To better understand each loss term's impact, we conducted a thorough ablation study of the global and local terms for both reconstructions and global-guided consistency learning. 
Table~\rr{\ref{tab:ablation_study}} showcases the ablation studies conducted on the BTCV validation datasets under both linear and end-to-end segmentation settings, as well as on the MM-WHS under end-to-end segmentation. ViT-B and ViT-T were both considered as the backbone of the framework. The first row represents the supervised baseline without any related strategies. Instead of reconstructing the local patches  at a time in each iteration, in the 2nd row, the proposed GL-MAE method firstly reconstructs the local patches $q$ times each iteration, thereby learning rich representation and exhibiting better performance most of the time. In the 3rd row, the reconstruction for global patches was used, further improving performance since the model can learn the global context as well as the local details. In the 4th row, the global-to-global consistency was introduced to learn a more robust representation of distortion caused by masking and learn the critical information, leading to further performance improvement. The last row represents the global-local consistency, which aims to capture the relationship between different parts of the images and their main semantics. The final objective loss function achieved the best performance across several datasets with various settings, demonstrating the importance of global information for volumetric data.

\begin{figure}[ht!]
	\centering
    \vspace{-5pt}
	\includegraphics[width=0.49\linewidth, height=3.4cm]{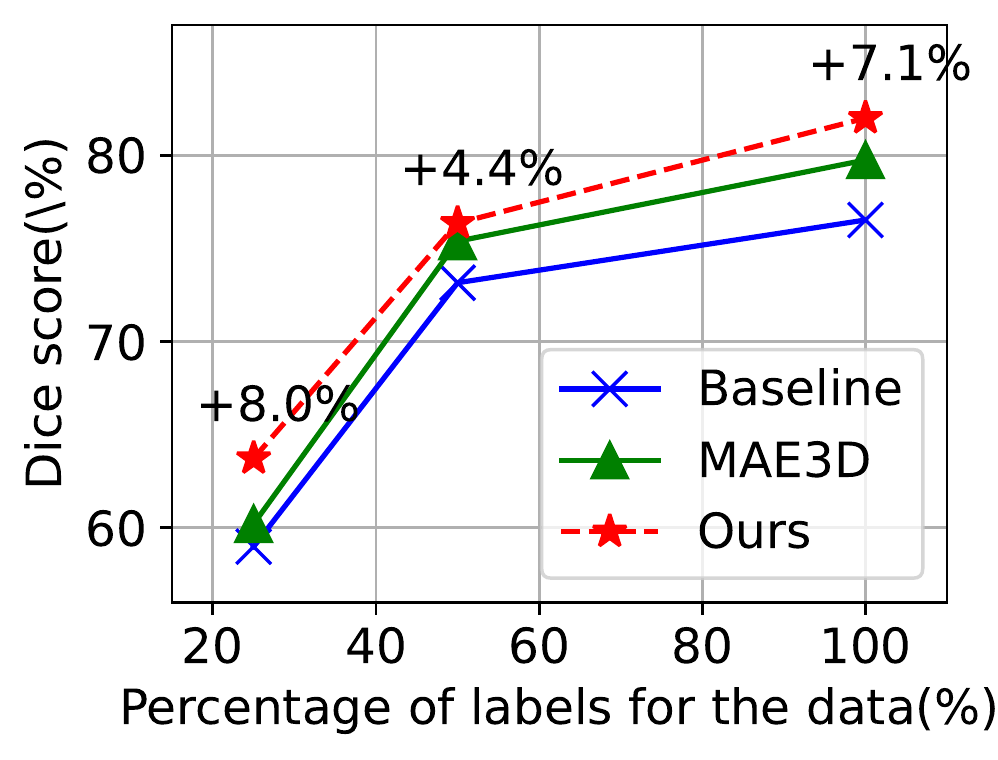}\hfill
	\includegraphics[width=0.49\linewidth, height=3.4cm]{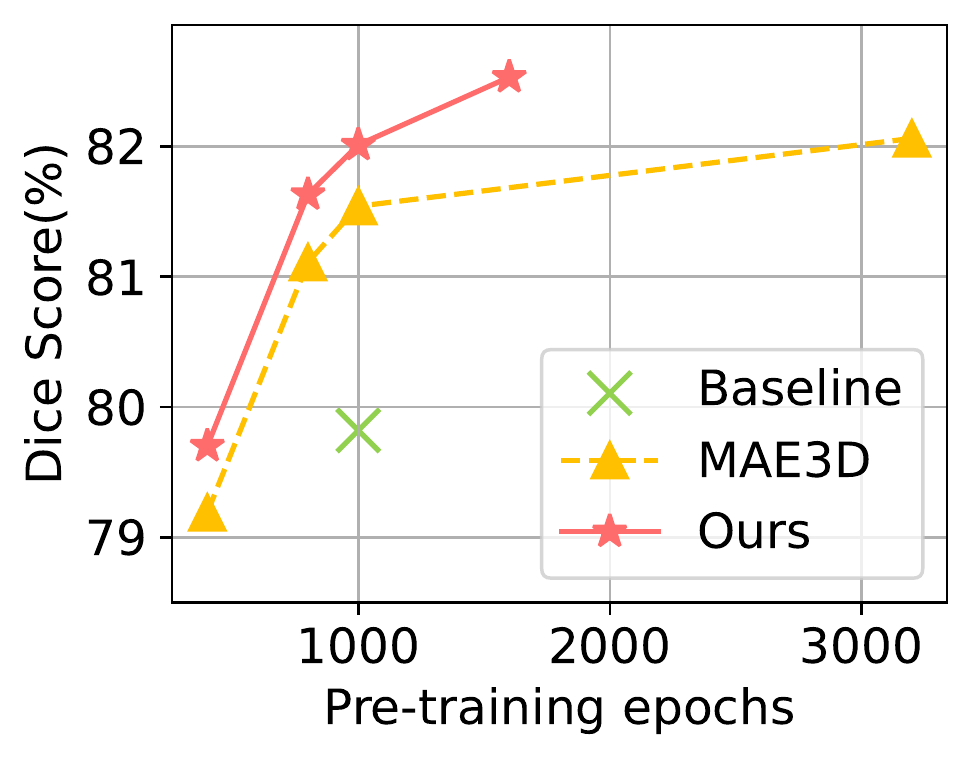}
    \vspace{-8pt}
	\caption{Left figure: Dice score(\%) of Linear evaluation with 25\%, 50\% and 100\% annotated training data on BTCV validation dataset. Right figure: Dice score(\%) of MAE3D and GL-MAE when the pre-training epochs increases.}
    \label{fig:semi_supervised_learning}
	\label{fig:training_epochs}
    \vspace{-13pt}
\end{figure}


\noindent\textbf{Label-efficient finetuning.}\label{sec:semiSupervised}
Following~\cite{tang2022self}, we conducted an evaluation of GL-MAE under a semi-supervised learning scheme with ViT-T as the backbone. 
The experimental results, as shown in Figure~\rr{\ref{fig:semi_supervised_learning}}, suggest that our proposed method can improve the dice score even when the amount of annotated training data is limited, with a 4.4\% to 8.0\% improvement. Transformer-based models are prone to over-fitting the limited labeled data due to their dense connections. However, our findings indicate that GL-MAE can reduce the necessity for labeled data and effectively enhance performance even in low-annotated learning scenarios.

\noindent\textbf{Convergence comparison.} 
As shown in Figure~\rr{\ref{fig:training_epochs}}, GL-MAE exhibits faster convergence and superior performance compared to MAE3D. This suggests that pre-training with global complete views and masked views can help stabilize the training process, resulting in faster convergence and more powerful representation. By utilizing global complete views as an 'anchor' for local and global masked views, the model establishes a stronger relationship through global-guided consistency. The integration of global context information and different scale reconstruction enhances the overall performance of GL-MAE and contributes to its superior results.

\noindent\textbf{Analysis of mask ratio.} The table in the Section Experiment of the supplementary material analyzed the impact of the mask ratio for the reconstruction of GL-MAE, and the findings showed that a mask ratio of 0.6 would be best. The results indicated that a very high mask ratio, like 0.7, is not ideal for this type of reconstruction, which is consistent with the findings on MAE3D in our implementation. Thus, it's crucial to carefully choose the mask ratio to balance between preserving important information and providing enough diversity for the model to learn robust representations.

\noindent\textbf{Scaling to larger data.} 
Pre-training methods should be able to scale to a large scale of data and demonstrate better performance~\cite{singh2023effectiveness}. 
As shown in the table in Section Experiment in the supplementary material, GL-MAE achieved a better dice score from 79.70\% to 81.41\% on the BTCV validation dataset with the same 400 epochs of pre-training. These results demonstrate the ability of GL-MAE to scale to larger amounts of data and improve performance on downstream tasks.


\noindent\textbf{Visualization.} GL-MAE was found to improve the completeness of segmentation results, as shown in Figure~\rr{\ref{fig:SegVisualizatoin}}. The results of segmentation using GL-MAE were better than those using MAE3D and Swin-UNETR, particularly in terms of completeness for larger organs.
 
 \begin{figure}[t]
   \vspace{-5pt}
    \centering
     \includegraphics[width=1.0\linewidth]{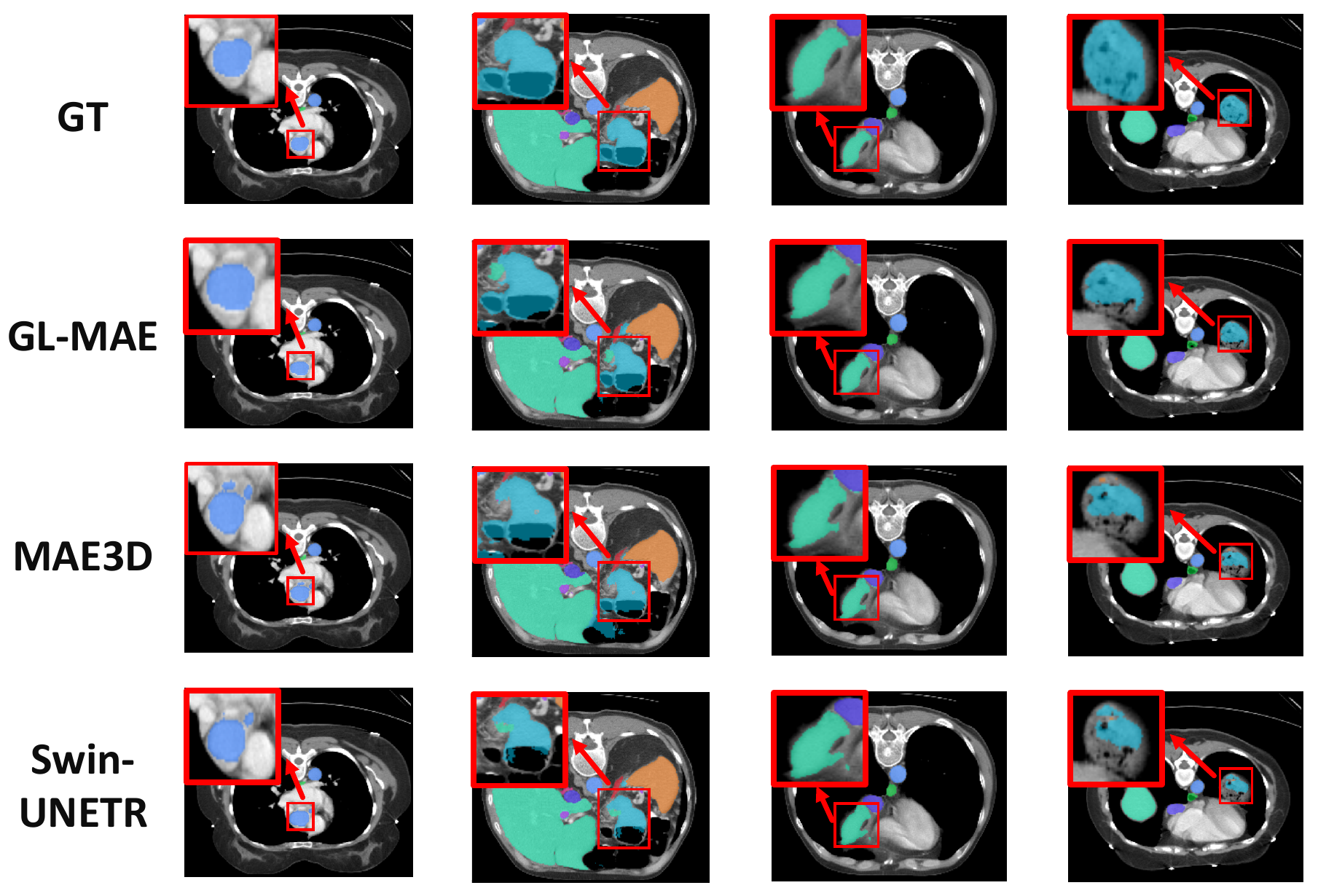}
     \vspace{-12pt}
     \caption{Visualization of segmentation results on BTCV validation dataset. The label blended with volumetric data (1st row), GL-MAE's
     prediction results (2nd row). Best viewed by zooming on visualization. More visualization results can be found in the supplementary material.}
    \vspace{-15pt}
     \label{fig:SegVisualizatoin}
 \end{figure}

\section{Conclusion}\label{sec:conclusion}


This paper proposes GL-MAE, a simple yet effective SSL pre-training strategy for volumetric images in medical image analysis.
We facilitated SSL with global information, which was neglected by previous ViT-based pre-training methods for volumetric image analysis, via the proposed global-local reconstruction and global-guided consistency learning.
The proposed method outperformed other state-of-the-art SSL algorithms on multiple downstream datasets, demonstrating its effectiveness and generalizability as a promising pre-training backbone for dense prediction on volumetric medical images.
Our future work will further investigate the scalability of the proposed GL-MAE with larger datasets and larger model capacity.


\bibliography{egbib,egbib_2}

\end{document}